\documentclass[runningheads]{llncs}
\usepackage{graphicx}
\usepackage{amsmath,amssymb} 
\usepackage{color}
\usepackage[width=122mm,left=12mm,paperwidth=146mm,height=193mm,top=12mm,paperheight=217mm]{geometry}

\usepackage{makecell}
\usepackage{booktabs}
\usepackage{multirow}
\usepackage{siunitx}

\usepackage[T1]{fontenc}
\usepackage{fix-cm}

\newcolumntype{?}{!{\vrule width 1pt}}
\newcolumntype{C}[1]{>{\centering}m{#1}}

\newcolumntype{X}{@{\hskip\tabcolsep\vrule width 1.5pt\hskip\tabcolsep}}

\usepackage{wrapfig,lipsum,booktabs}
\usepackage{caption}
\usepackage{pifont}
\newcommand{\cmark}{\ding{51}}%
\newcommand{\xmark}{\ding{55}}%

\begin{document}
\pagestyle{headings}
\mainmatter

\title{Object Detection in Video with\\Spatiotemporal Sampling Networks} 

\titlerunning{Object Detection in Video with Spatiotemporal Sampling Networks}

%
%

\author{Gedas Bertasius$^{1}$, Lorenzo Torresani$^{2}$, and Jianbo Shi$^{1}$\\
$^{1}$University of Pennsylvania, $^{2}$Dartmouth College}
\titlerunning{Object Detection in Video with Spatiotemporal Sampling Networks}
\authorrunning{Gedas Bertasius, LorenzoTorresani, and Jianbo Shi}
\institute{}

\authorrunning{G. Bertasius, L. Torresani, J. Shi}

\maketitle

\begin{abstract}

We propose a Spatiotemporal Sampling Network (STSN) that uses deformable convolutions across time for object detection in videos. Our STSN performs object detection in a video frame by learning to spatially sample features from the adjacent frames. This naturally renders the approach robust to occlusion or motion blur in individual frames. Our framework does not require additional supervision, as it optimizes sampling locations directly with respect to object detection performance. Our STSN outperforms the state-of-the-art on the ImageNet VID dataset and compared to prior video object detection methods it uses a simpler design, and does not require optical flow data for training.
\end{abstract}

\section{Introduction}

\begin{figure}[t]
\begin{center}
   \includegraphics[width=1\linewidth]{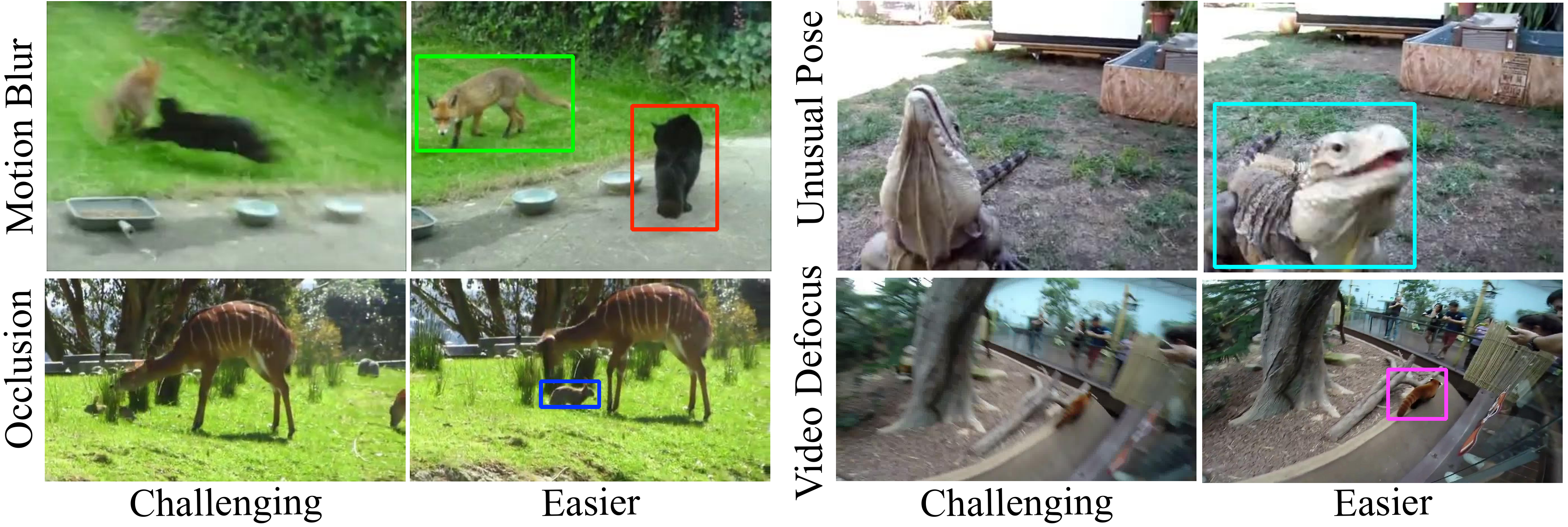}
\end{center}
\vspace{-0.4cm}
   \caption{An illustration of the common challenges associated with object detection in video. These include video defocus, motion blur, occlusions and unusual poses. The bounding boxes denote the objects that we want to detect in these examples.\vspace{-0.5cm}}
\label{intro_fig}
\end{figure}

In recent years, deep convolutional networks have achieved remarkable results in many computer vision tasks~\cite{NIPS2012_4824,43022,Simonyan14c,gberta_2015_CVPR,He2016DeepRL,xie2016groups,DBLP:journals/corr/ToshevS13,gberta_2017_CVPR}, including object detection in images~\cite{SPP,he2017maskrcnn,lin2017focal,ren2015faster,girshick15fastrcnn,girshick2014rcnn,guptaECCV14,44872,dai16rfcn,DBLP:journals/corr/RedmonDGF15,DBLP:journals/corr/RedmonF16}. However, directly applying these image-level models to object detection in video is difficult due to motion blur, video defocus, unusual poses, or object occlusions (see Figure~\ref{intro_fig}). Despite these challenges, it is natural to assume that video object detectors should be more powerful than still image detectors because video contains richer information about the same object instance (e.g., its appearance in different poses, and from different viewpoints). The key challenge then is designing a model that effectively exploits temporal information in videos.

Prior work~\cite{DBLP:journals/corr/KangLYZYXZWWWO16,DBLP:journals/corr/KangOLW16,DBLP:journals/corr/HanKPRBSLYH16,DBLP:journals/corr/LeeEJR16} has proposed to exploit such temporal information in videos by means of various post-processing steps aimed at making object detections coherent across time. However, since temporal coherence is enforced in a second stage, typically these methods cannot be trained end-to-end. To overcome this limitation, recent work~\cite{zhu17fgfa} has introduced a flow-based aggregation network that is trainable end-to-end. It exploits optical flow to find correspondences across time and it then aggregates features across temporal correspondences to smooth object detections over adjacent frames. However, one of the downsides of this new model is that in addition to performing object detection, it also needs to predict motion. This is disadvantageous due to the following reasons: 1) designing an effective flow network architecture is not trivial, 2) training such a model requires large amounts of flow data, which may be difficult and costly to obtain, 3) integrating a flow network and a detection network into a single model may be challenging due to factors such as different loss functions, differing training procedures for each network, etc.
 
 
To address these shortcomings, in this work, we introduce a simple, yet effective Spatiotemporal Sampling Network (STSN) that uses deformable convolutions~\cite{8237351} across space and time to leverage temporal information for object detection in video. Our STSN learns to spatially sample useful feature points from nearby video frames such that object detection accuracy in a given video frame is maximized. To achieve this, we train our STSN end-to-end on a large set of video frames labeled with bounding boxes. We show that this leads to a better accuracy compared to the state-of-the-art on the ImageNet VID dataset~\cite{ILSVRC15}, without requiring complex flow network design, or the need to train the network on large amounts of flow data. 

\captionsetup{labelformat=default}
\captionsetup[figure]{skip=10pt}

\section{Related Work}


\subsection{Object Detection in Images}

Modern object detectors~\cite{SPP,he2017maskrcnn,lin2017focal,ren2015faster,girshick15fastrcnn,girshick2014rcnn,guptaECCV14,44872,dai16rfcn,DBLP:journals/corr/RedmonDGF15,DBLP:journals/corr/RedmonF16} are predominantly built on some form of deep CNNs~\cite{NIPS2012_4824,Simonyan14c,He2016DeepRL}. One of the earliest deep CNN object detection systems was R-CNN~\cite{girshick2014rcnn}, which involved a two-stage pipeline where object proposals were extracted in the first stage, and then each proposal was classified using a CNN. To reduce the computational burden, the methods in~\cite{SPP}, and~\cite{girshick15fastrcnn} leveraged ROI pooling, which led to more efficient learning. Furthermore, to unify the entire object detection pipeline, Faster R-CNN~\cite{ren2015faster} replaced various region proposal methods by another network to make the entire system trainable end-to-end. Following this work, several methods~\cite{DBLP:journals/corr/RedmonDGF15,DBLP:journals/corr/RedmonF16} extended Faster R-CNN into a system that runs in real time with small reduction in performance.  Additionally, recent work~\cite{dai16rfcn} introduced position sensitive ROI pooling, which significantly improved the detection efficiency compared to prior object detection systems.  Finally,  two recent methods, Mask R-CNN~\cite{he2017maskrcnn}, and Deformable CNNs~\cite{8237351}, improved object detection results even further and they represent the current state-of-the-art in object detection. Whereas Mask-RCNNs use an additional branch that predicts a mask for each region of interest, Deformable CNNs employ deformable convolutions, which allow the network to condition discriminatively its receptive field on the input, and to also model deformations of objects more robustly.

While the aforementioned methods work well on images, they are not designed to exploit temporal relationships in video. Instead, our Spatiotemporal Sampling Network (STSN), is specifically designed for a video object detection task. Unlike standard Deformable CNNs~\cite{8237351}, which use deformable convolution in the spatial domain, our STSN learns to sample features temporally across different video frames, which leads to improved video object detection accuracy.


\subsection{Object Detection in Videos}


Up until the introduction of the ImageNet VID challenge~\cite{ILSVRC15}, there were no large-scale benchmarks for video object detection. Thus, there are only few methods that we can compare our work to. T-CNNs~\cite{DBLP:journals/corr/KangLYZYXZWWWO16,DBLP:journals/corr/KangOLW16} use a video object detection pipeline that involves predicting optical flow first, then propagating image-level predictions according to the flow, and finally using a tracking algorithm to select temporally consistent high confidence detections. Seq-NMS~\cite{DBLP:journals/corr/HanKPRBSLYH16} constructs a temporal graph from overlapping bounding box detections across the adjacent frames, and then uses dynamic programming to select bounding box sequences with the highest overall detection score. The work of Lee et al.~\cite{DBLP:journals/corr/LeeEJR16} treats a video object detection task as a multi-object tracking problem. Finally, the method of Feichtenhofer et al.~\cite{feichtenhofer2017detect} proposes a ConvNet architecture that solves detection and tracking problems jointly, and then applies a Viterbi algorithm to link the detections across time.

The approach most similar to our work is the method of Zhu et al.~\cite{zhu17fgfa}, who proposed an end-to-end trainable network that jointly estimates optical flow and also detects objects in video. This is accomplished by using the predicted optical flow to align the features from the adjacent frames. The aggregated features are then fed as input to the detection network. 

Our method is beneficial over the methods that use optical flow CNNs such as the method of Zhu et al.~\cite{zhu17fgfa}. First, we note that pretrained optical flow CNNs do not always generalize to new datasets, which may hinder video object detection performance. In contrast, our method has a learnable spatiotemporal sampling module that is discriminatively trained from object detection labels, and thus, it does not suffer from this issue. Furthermore, our STSN can be trained for video object detection in a single stage end-to-end. In comparison, methods that rely on optical flow require an additional stage to train an optical flow CNN, which renders the training procedure more cumbersome and lengthy.  For example, we note that it would take about four days to train an optical flow CNN of FGFA~\cite{zhu17fgfa} from scratch and then four additional days to train FGFA~\cite{zhu17fgfa}  for video object detection, making it eight days of total training time. In contrast, our STSN is trained in a single stage in only 4 days. Finally, we point out that our STSN also yields a gain ---albeit moderate--- in video object detection accuracy.

\section{Background: Deformable Convolution}

Before describing our method, we first review some background information on deformable convolution~\cite{8237351}, which is one of the key components of our STSN. Let us first note that a standard 2D convolution is comprised of two steps: 1) sampling locations on a uniformly-spaced grid $\mathcal{R}$, and 2) performing a weighted summation of sampled values using weights $w$. For example, if we consider a standard 2D convolution with a $3 \times 3$ kernel, and a dilation factor of $1$, the grid $\mathcal{R}$ is defined as $\mathcal{R} = \{(-1,-1),(-1,0), \hdots, (0,1),(1,1)\}$. Under a standard 2D convolution, to compute a new value at pixel location $p_0$ in the output feature map $y$, we would perform the following operation on the input feature map $x$:

\vspace{-0.4cm}
\begin{equation}
\begin{split}
y(p_0) =\sum_{p_n \in \mathcal{R}} w(p_n) \cdot x(p_0 + p_n),
\end{split} 
\end{equation}
\vspace{-0.4cm}

Instead, in a deformable 2D convolution, the grid $\mathcal{R}$ is augmented with data-conditioned offsets $\{\Delta p_n | n=1, \hdots, N\}$, where $N=|\mathcal{R}|$. We can then compute a deformable convolution as: 

\vspace{-0.4cm}
\begin{equation}
\begin{split}
y(p_0) =\sum_{p_n \in \mathcal{R}} w(p_n) \cdot x(p_0 + p_n + \Delta p_n)
\end{split} 
\end{equation}
\vspace{-0.4cm}

Since the offset $\Delta p_n$ is typically fractional, the operation above is implemented using bilinear interpolation. Note that the offsets are obtained by applying a separate convolutional layer to the  activation tensor containing the feature map $x$. This yields an offset map that has the same spatial resolution as the input feature map. Also, note that the offsets are shared across all feature channels of a given activation tensor. During training, the weights for the deformable convolution kernel, and the offsets kernel are learned jointly by propagating gradients through the bilinear interpolation operator. We refer the reader to the original work that introduced deformable convolutions~\cite{8237351} for further details.

\captionsetup{labelformat=default}
\captionsetup[figure]{skip=10pt}

\begin{figure}[t]
\begin{center}
   \includegraphics[width=0.88\linewidth]{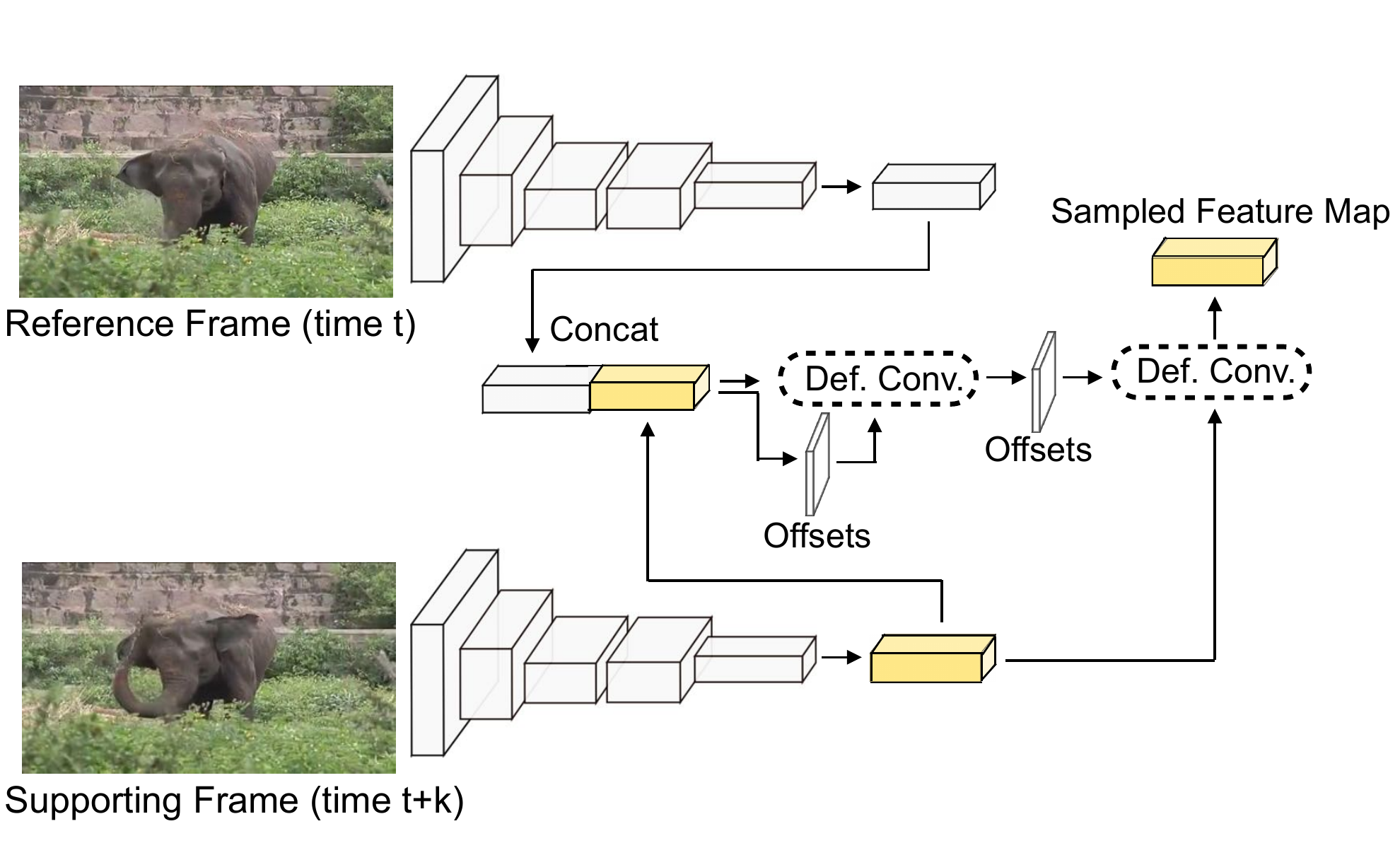}
\end{center}
\vspace{-0.7cm}
        \caption{Our spatiotemporal sampling mechanism, which we use for video object detection. Given the task of detecting objects in a particular video frame (i.e., a reference frame), our goal is to incorporate  information from a nearby frame of the same video (i.e., a supporting frame). First, we extract features from both frames via a backbone convolutional network (CNN). Next, we concatenate the features from the reference and supporting frames, and feed them through multiple deformable convolutional layers. The last of such layers produces offsets that are used to sample informative features from the supporting frame. Our spatiotemporal sampling scheme allows us to produce accurate detections even if objects in the reference frame appear blurry or occluded.\vspace{-0.5cm}} 
\label{arch1_fig}
\end{figure}

\section{Spatiotemporal Sampling Network}

Our goal is to design a network architecture that incorporates temporal information for object detection in video. 

Let us denote with $I_t$ the frame at time $t$ in the video. Let us consider one of the scenarios depicted in Figure~\ref{intro_fig}, e.g., a setting where $I_t$ is blurry, contains an object in an unusual pose, or perhaps an occlusion. But let us assume that a nearby frame $I_{t+k}$ includes the same object clearly visible and in a relatively standard pose. If we only had access to $I_{t}$, accurate object detection would be very challenging. However, leveraging information from $I_{t+k}$ may enable more robust detection in the frame $I_{t}$ . Thus, the main challenge in this setting is incorporating object-level information from the {\em supporting} frame $I_{t+k}$ for an improved object detection accuracy in the {\em reference} frame $I_t$. Note that in our system each frame in the video is treated in turn as a reference frame in order to produce object detection in every frame of the video. Furthermore, in practice we use $2 K$ supporting frames for detection in the reference frame, by taking the $K$ preceding frames and the $K$ subsequent frames as supporting frames, i.e. $\{I_{t-K}, I_{t-(K-1)}, \hdots, I_{t-1}, I_{t+1}, \hdots, I_{t+(K-1)}, I_{t+K}\}$. However, for ease of explanation we introduce our STSN by considering a single supporting frame $I_{t+k}$.

To effectively integrate temporal information we need two things: 1) powerful object-level features from an image-level network, and 2) an ability to sample useful object-level features from the supporting frames for the reference frame. We achieve the former by employing a state-of-the-art backbone network. For the latter, we design a spatiotemporal sampling scheme, which we describe below. 

Our STSN can be summarized in four steps. First, a backbone convolutional network computes object-level features for each video frame individually. Then, spatiotemporal sampling blocks are applied to the object-level feature maps in order to sample relevant features from nearby frames conditioned on the input {\em reference} frame.  Next, the sampled features from each video frame are temporally aggregated into a single feature tensor for the reference frame using a per-pixel weighted summation. Finally, the  feature tensor is provided as input to the detection network to produce final object detection results for the given reference frame.  We note that our framework integrates these conceptually-distinct four steps into a single architecture, which we train end-to-end. 


\textbf{Backbone Architecture.} Our backbone network is applied to each frame of the video. As backbone network, we use a Deformable CNN~\cite{8237351} based on the ResNet-101~\cite{He2016DeepRL} architecture, which is one of the top-performing object detection systems at the moment. Similarly to~\cite{8237351}, our backbone network employs $6$ deformable convolutional layers. We also note that even though we use a Deformable CNN architecture, our system can easily integrate other architectures and thus it can benefit from future improvements in still-image object detection. 



\textbf{Spatiotemporal Feature Sampling.} Our main contribution is the design of a spatiotemporal sampling mechanism, which seamlessly integrates temporal information in a given video. As a first step, we feed the reference frame $I_t$ and the supporting frame $I_{t+k}$ through our image-level backbone network, which produces feature tensors $f_t$ and  $f_{t+k}$, respectively. Note that $f_t, f_{t+k} \in \mathbb{R}^{c \times h \times w}$ where $c, h$, and $w$ are the number of channels, the height, and the width of the activation tensor. The feature tensors $f_t$, and $f_{t+k}$ are then concatenated into a new feature tensor $f_{t,t+k} \in \mathbb{R}^{2c \times h \times w}$. Note that this tensor $f_{t,t+k}$ now has twice as many channels as our initial tensors, and that it now contains object-level information from both the reference and the supporting frame. 


Next, we use the tensor $f_{t,t+k}$ to predict $(x,y)$ location offsets, which are then used to sample the supporting tensor $f_{t+k}$. The sampling mechanism is implemented using a deformable convolutional layer, which takes 1) the predicted offsets, and 2) the supporting tensor $f_{t+k}$ as its inputs, and then outputs a newly sampled feature tensor $g_{t,t+k}$, which can be used for object detection in the reference frame.  We use subscript ${t,t+k}$ to denote the resampled tensor because, although $g$ is obtained by resampling the {\em supporting} tensor, the offset computation uses both the reference as well as the supporting frame. A detailed illustration of our spatiotemporal sampling scheme is presented in Figure~\ref{arch1_fig}. 

In practice, our spatiotemporal sampling block has $4$ deformable convolution layers (only $2$ are shown in Figure~\ref{arch1_fig}). This means that the initially predicted offsets \smash{$o^{(1)}_{t,t+k}$} and the concatenated temporal features $f_{t,t+k}$ are first used as inputs to a deformable convolution layer that outputs a new feature map \smash{$g^{(1)}_{t,t+k}$}. Next, we use \smash{$g^{(1)}_{t,t+k}$} to predict offsets \smash{$o^{(2)}_{t,t+k}$}, and a new feature map \smash{$g^{(2)}_{t,t+k}$}. 
This continues for 2 more layers until we obtain offsets \smash{$o^{(4)}_{t,t+k}$}, which are then used to sample the points out of the supporting feature map $f_{t+k}$. The final sampled feature map \smash{$g^{(4)}_{t,t+k}$} is obtained via another deformable convolutional layer that takes as inputs offsets \smash{$o^{(4)}_{t,t+k}$} and the original supporting feature map $f_{t+k}$. 

Our proposed spatiotemporal sampling mechanism learns, which object-level features in the supporting frame are useful for object detection in the reference frame. Conceptually, it replaces the optical flow used in~\cite{zhu17fgfa} to establish temporal correspondences with a learnable module that is discriminatively trained from object detection labels. In our experimental section, we show that such a sampling scheme allows us to improve video object detection performance over the still-image baseline and the flow-based method of Zhu et al.~\cite{zhu17fgfa} without training our model on optical flow data.

\textbf{Feature Aggregation.} The spatiotemporal sampling procedure is applied for all the supporting frames in the selected range. Note, that this  includes a special case, when the reference frame is treated as a supporting frame to itself to produce $g^{(4)}_{t,t}$, which is a feature tensor computed from only the reference frame. 

The resulting feature tensors have the following form: \smash{$g^{(4)}_{t,t+k} \in \mathbb{R}^{c^{(4)} \times h \times w}$}. These feature tensors are aggregated into an output feature tensor \smash{$g^{agg}_{t} \in \mathbb{R}^{c^{(4)} \times h \times w}$} for the reference frame. This tensor captures information from the reference frame, its $K$ preceding frames and its $K$ subsequent frames.  The output tensor value $g^{agg}_t(p)$ for frame $t$ at pixel $p$ is computed as a weighted summation:

\vspace{-0.5cm}
\begin{equation}
\begin{split}
g^{agg}_t(p)= \sum_{k=-K}^{K} w_{t,t+k}(p) ~g^{(4)}_{t,t+k}(p)
\end{split}
\end{equation}
\vspace{-0.4cm}

Inspired by strong results presented in~\cite{zhu17fgfa}, we use their proposed feature aggregation method where the weights $w$ indicate the importance of each supporting frame to the reference frame. To compute the weights $w$, we attach a  3-layer subnetwork $S(x)$ to the features $g^{(4)}_{t,t+k}$ and then compute their intermediate feature representations $S(g^{(4)}_{t,t+k})$. We then obtain the weights $w$ by applying an exponential function on the cosine similarity between each corresponding feature point in a reference frame and a supporting frame:

\vspace{-0.3cm}
\begin{equation}
\begin{split}
w_{t,t+k}(p) = \exp{\left(\frac{S(g^{(4)}_{t,t})(p) \cdot S(g^{(4)}_{t,t+k})(p) }{ |S(g^{(4)}_{t,t})(p)||S(g^{(4)}_{t,t+k})(p)|}\right)}
\end{split}
\end{equation}
\vspace{-0.3cm}

Finally, all weights $w$ are fed into the softmax layer, to ensure that the weights sum up to $1$ at each pixel location $p$ (i.e., $\sum_{k=-K}^{K} w_{t,t+k}(p) = 1 \hspace{0.1cm} \forall p$). 

\textbf{Object Detection.} Finally, the aggregated feature tensor $g^{agg}_t$ is used as input to the detection network, which outputs the final bounding box predictions and their object class probabilities. We describe more details related to the detection network in the next section along with other implementation details.

\subsection{Implementation Details}

For our experiments we use the MXNet~\cite{journals/corr/ChenLLLWWXXZZ15} library. Below we provide details related to our STSN architecture, and our training and inference procedures.



\textbf{Architecture.} For our backbone network we adopt a state-of-the-art Deformable CNN~\cite{8237351} based on the ResNet-101~\cite{He2016DeepRL} architecture. Our spatiotemporal sampling block consists of four $3 \times 3$ deformable convolutional layers each with $1024$ output channels. In addition, it also has four $3 \times 3$ convolutional layers predicting $(x,y)$ offsets. To implement a subnetwork $S(x)$ that predicts feature aggregation weights, we use a sequence of $1 \times 1$, $3 \times 3$ and $1 \times 1$  convolutional layers with $512, 512$ and $2048$ output channels respectively. Our detection network is implemented based on the deformable R-FCN design~\cite{dai16rfcn,zhu17dff,8237351}.  When feeding the aggregated feature $g^{agg}_t$ to the detection network, we split its $1024$ channels into two parts, and feed the first and the last $512$ channels to the RPN and R-FCN sub-networks respectively. For the RPN, we use $9$ anchors and $300$ proposals for each image. Furthermore, for the R-FCN, we use deformable position-sensitive ROI pooling with $7\times7$ groups.



\textbf{Training.} Our entire STSN model is fully differentiable, and thus, trainable end-to-end. During training, we resize all input images to a shorter side of $600$ pixels, and use $T=3$ frames to train our model (i.e., $K=1$). More specifically, we randomly sample one supporting frame before and one supporting frame after the reference frame. We observed that using more supporting frames in training does not lead to a higher accuracy. 

For the rest of our training procedure, we follow the protocol outlined in~\cite{zhu17fgfa}. Specifically, we train our model in two stages. First, we pre-train our full model on the Imagenet DET dataset using the annotations of the $30$ object classes that overlap with the Imagenet VID dataset. Note that Imagenet DET dataset  contains only images, and thus, we cannot sample meaningful supporting frames in this case. Therefore, in the case of images, we use the reference frames as our supporting frames.  Afterwards, the entire model is trained for $120K$ iterations on $4$ Tesla K40 GPUs with each GPU holding a single mini-batch. The learning rate is set to $0.001$ and $0.0001$ for the first $80K$ and the last $40K$ iterations respectively.  Afterwards, we finetune the entire model on the Imagenet VID dataset for $60K$ iterations with a learning rate of $0.001$ and $0.0001$ for the first $40K$ and the last $20K$ iterations respectively. Note that in the second stage of training we sample the supporting frames randomly within a certain neighborhood of a reference frame (as described above).



\textbf{Inference.} During inference, we use $T=27$, meaning that we consider $K=13$ supporting frames before and after the reference frame. To avoid GPU memory issues, we first extract features from the backbone network for each image individually, and then cache these features in the memory. Afterwards, we feed all these features into our spatiotemporal sampling block. At the end, standard NMS with a threshold of $0.3$ is applied to refine the detections. To handle the first and the last $K=13$ frames in the video ---two boundary cases that require sampling the neighboring frames beyond the video start and end, we pad the start of a video with $K$ copies of the first frame, and the end of a video with $K$ copies of the last frame.

\section{Experimental Results}

In this section, we evaluate our approach for video object detection on the ImageNet VID~\cite{ILSVRC15} dataset, which has $3,862$ and $555$ training and testing video clips respectively. Each video is annotated with bounding boxes. The frames from each video are extracted at $25-30$ fps. The dataset contains $30$ object categories that are a subset of the $200$ categories in the ImageNet DET dataset.

   \setlength{\tabcolsep}{1.25pt}

     \begin{table}[t]
   \scriptsize
    \begin{center}
    \begin{tabular}{ c | c | c | c | c | c | c |}
    \cline{2-7}
    & \multicolumn{6}{ c |}{Methods} \\
    \cline{2-7}
     &  \multicolumn{1}{ c |}{D\&T~\cite{feichtenhofer2017detect}}  &   \multicolumn{1}{ c |}{Our SSN} & \multicolumn{1}{ c |}{FGFA~\cite{zhu17fgfa}}  &  \multicolumn{1}{ c |}{Our STSN} & \multicolumn{1}{ c |}{D\&T+~\cite{feichtenhofer2017detect}}  &  \multicolumn{1}{ c |}{Our STSN+}\\ \cline{1-7}
     \multicolumn{1}{| c |}{No FlowNet?} &  \cmark & - & \xmark & \cmark & \cmark & \cmark\\ \hline
      \multicolumn{1}{| c |}{Not Using Flow Data?} & \cmark  & -  &  \xmark & \cmark & \cmark & \cmark\\ \hline
       \multicolumn{1}{| c |}{No Temporal Post-Processing?} & \cmark & - &  \cmark & \cmark & \xmark & \xmark\\ \hline
      \multicolumn{1}{| c |}{mAP@0.5} & 75.8 & 76.0 & 78.8 & 78.9 & 79.8 & \bf 80.4\\ \hline
    \end{tabular}
    \end{center}
    \footnotesize
      \caption{We use the ImageNet VID~\cite{ILSVRC15} dataset to compare our STSN to the state-of-the-art FGFA~\cite{zhu17fgfa} and D\&T~\cite{feichtenhofer2017detect} methods. Note that SSN refers to our static baseline, which is obtained by using only the reference frame for output generation (no temporal info). Also note, that D\&T+ and STSN+ refer to D\&T and STSN baselines with temporal post-processing applied on top of the CNN outputs. Based on these results, we first point out that unlike FGFA, our STSN does not rely on the external optical flow data, and still yields higher mAP ($\bf 78.9$ vs $\bf 78.8$). Furthermore, when no temporal post-processing is used, our STSN produces superior performance  in comparison to the D\&T baseline ($\bf 78.9$ vs $\bf 75.8$).  Finally, we demonstrate that if we use a simple Seq-NMS~\cite{DBLP:journals/corr/HanKPRBSLYH16} temporal post-processing scheme on top of our STSN predictions, we can further improve our results and outperform all the other baselines.\vspace{-0.7cm}}  
    \label{results1_table}
   \end{table}


\subsection{Quantitative Results} 

To assess the effectiveness of our method we compare it to several relevant baselines, mainly two state-of-the-art methods FGFA~\cite{zhu17fgfa} and D\&T~\cite{feichtenhofer2017detect}. First, to verify that using temporal information from video is beneficial, we include a static image-level variant of our model (SSN) that uses only the reference frame to make its predictions. Furthermore, we also want to show that our spatiotemporal sampling scheme works as effectively as the optical flow network in~\cite{zhu17fgfa}, but without requiring optical flow supervision. To do so, we replace the optical flow network from~\cite{zhu17fgfa}, with our spatiotemporal sampling mechanism. The rest of the architecture and the training details are kept the same for both baselines. Such an experimental design allows us to directly compare the effectiveness of our spatiotemporal sampling scheme and the optical flow network of FGFA~\cite{zhu17fgfa}. 

Finally, we demonstrate that our method performs better than the D\&T~\cite{feichtenhofer2017detect} method in two scenarios: 1) when we only use CNN-level outputs for video object detection, and also 2) when we allow temporal post-processing techniques such as Seq-NMS to be applied on top of the CNN outputs. We note that in Table~\ref{results1_table}, D\&T~\cite{feichtenhofer2017detect} and STSN refer to the CNN-level baselines whereas D\&T+~\cite{feichtenhofer2017detect} and STSN+ denote these same methods but with temporal post-processing (i.e. Seq-NMS~\cite{DBLP:journals/corr/HanKPRBSLYH16}, object-tube based linking~\cite{feichtenhofer2017detect}, etc)  applied on top of the CNN outputs.

We present our results in Table~\ref{results1_table},  where we assess each method according to several criteria. In the first row of  Table~\ref{results1_table}, we list whether a given method requires integrating a separate flow network into its training / prediction pipeline. Ideally, we would want to eliminate this step because optical flow prediction requires designing a highly complex flow network architecture. We also list whether a given method requires pre-training on the external optical flow data, which we would want to avoid since it makes the whole training pipeline more costly. Additionally, we list, whether a given method uses any external temporal post-processing steps, which we would want to eliminate because they typically make the training / prediction pipeline disjoint and more complex. Finally, we assess each method according to the standard mean average precision (mAP) metric at intersection over union (IoU) threshold of $0.5$.

  \captionsetup{labelformat=default}
\captionsetup[figure]{skip=10pt}
\begin{figure}[t]
\begin{center}
   \includegraphics[width=1\linewidth]{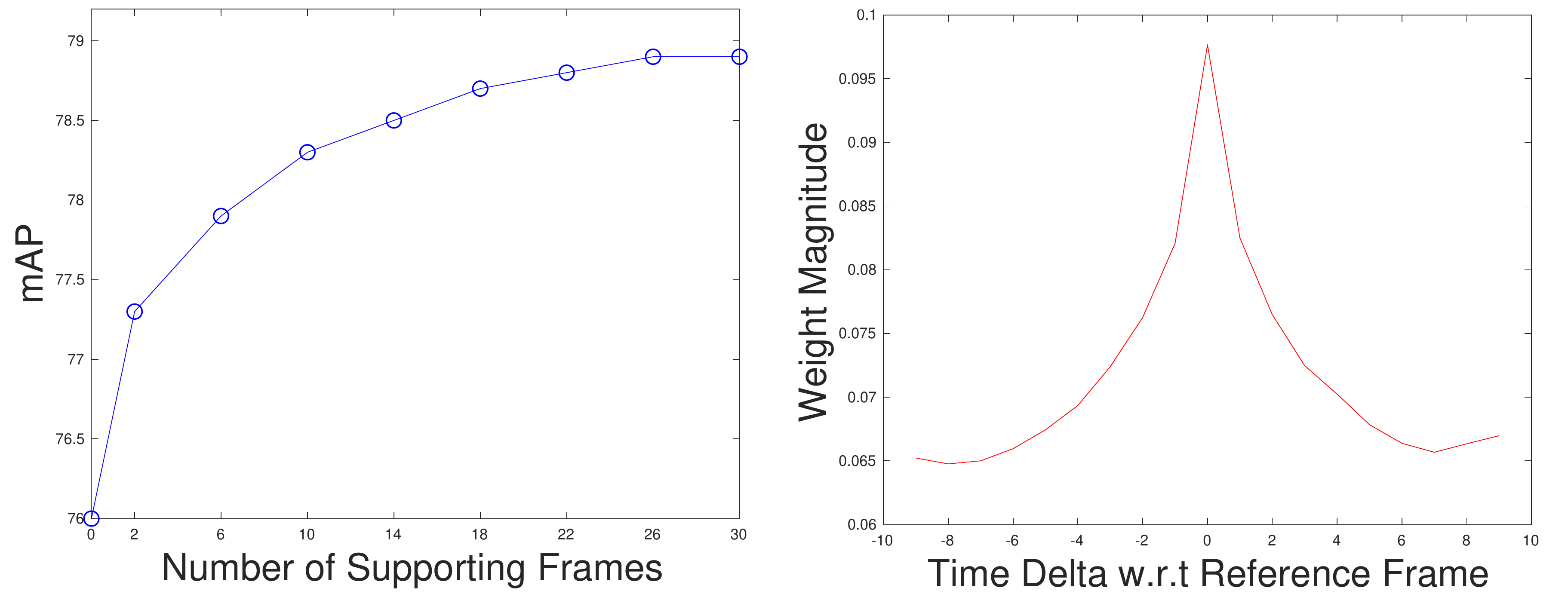}
\end{center}
\vspace{-0.4cm}
   \caption{A figure illustrating some of our ablation experiments. \textbf{Left:} we plot mAP as a function of the number of supporting frames used by our STSN. From this plot, we notice that the video object detection accuracy improves as we use more supporting frames. \textbf{Right:} To understand the contribution of each of the supporting frames, we plot the average weight magnitudes $w_{t,t+k}(p)$ for different values of $k$. Here, $p$ represents a point at the center of an object. From this plot, we observe that the largest weights are associated with the supporting frames that are near the reference frame. However, note that even supporting frames that are further away from the reference frame (e.g. $k=9$) contribute quite substantially to the final object detection predictions.} 
\label{results1_mAP_vs_frames}
\end{figure}

Based on our results in Table~\ref{results1_table}, we make the following conclusions. First, we note that our STSN produces better quantitative results than the state-of-the-art FGFA method ($\bf 78.9$ vs $\bf 78.8$). We acknowledge that our accuracy improvement over FGFA is moderate. However, we point out that our STSN operates in a much more challenging setting than FGFA. Unlike FGFA, our STSN does not use any optical flow supervision. Instead, it is trained directly for video object detection. The fact that STSN learns temporal correspondences without direct optical flow supervision, and still outperforms FGFA is quite impressive. Such results also show the benefit of discriminative end-to-end training with respect to the final video object detection task objective.

We next compare our STSN to the D\&T baseline~\cite{feichtenhofer2017detect}. We note that unlike for the FGFA~\cite{zhu17fgfa} baseline, it is much harder to make a direct comparison between STSN and D\&T. Whereas our STSN aims to produce powerful spatiotemporal features, the method of D\&T~\cite{feichtenhofer2017detect} is targeted more for smoothing the final bounding box predictions across time. Thus, we believe that these two methods are complementary, and it would be possible to integrate them together for the model that produces both: temporally smooth features, as well as temporally smooth bounding box predictions. We also note that our STSN and D\&T~\cite{feichtenhofer2017detect} use slightly different architectures (both based on ResNet-101 though). 


First, we compare STSN and D\&T in a setting when no temporal post-processing (i.e. Seq-NMS~\cite{DBLP:journals/corr/HanKPRBSLYH16}, object-tube linking~\cite{feichtenhofer2017detect}, etc)  is used, and show that our STSN outperforms the D\&T baseline by a substantial margin  ($\bf 78.9$ vs $\bf 75.8$). These results indicate, that our STSN is able to learn powerful spatiotemporal features, and produce solid video object detection results even without  temporal post-processing algorithms that link bounding box detections over time.

Afterwards, we show that integrating a simple temporal post-processing algorithm Seq-NMS~\cite{DBLP:journals/corr/HanKPRBSLYH16} further improves our STSN's results. Such a scheme allows us to outperform the D\&T+ baseline ($\bf 80.4$ vs $\bf 79.8$), which uses a similar Viterbi based temporal post-processing scheme.

\subsection{Ablation Studies}

\textbf{\indent{Optimal Number of Supporting Frames.}} In the left subplot of Figure~\ref{results1_mAP_vs_frames}, we also illustrate how the number of supporting frames affects the video object detection accuracy. We notice that the performance keeps increasing as we add more supporting frames, and then plateaus at $T=27$. 

\textbf{Increasing the Temporal Stride.} We also investigate how the temporal stride $k$, at which we sample the supporting frames, affects STSN's performance. We report that temporal strides of $k=2$ and $k=4$, yield mAP scores of $79.0$ and $77.9$, respectively. Thus, $k=2$ yields a slight improvement over our original $78.9$ mAP score. However, increasing $k$ to larger values reduces the accuracy.

\textbf{Feature Aggregation Weight Analysis.} To analyze how much each of the supporting frame contributes to the final object detections, we visualize the average weight magnitudes $w_{t,t+k}(p)$ for different values of $k$. This visualization is presented  in the right subplot of Figure~\ref{results1_mAP_vs_frames}. We note that in this case, the weight magnitudes correspond to the point $p$, which is located at the center of an object. From this plot, we can conclude that the largest contribution comes from the supporting frames that are near the reference frame ($k=-1, 0, 1)$. However, note that even even supporting frames that are further away from the reference frame (e.g. $k=-9, 9$) have non-zero weights, and contribute quite substantially to the final object detection predictions.


  \captionsetup{labelformat=default}
\captionsetup[figure]{skip=10pt}

\begin{figure}[t]
\begin{center}
   \includegraphics[width=1\linewidth]{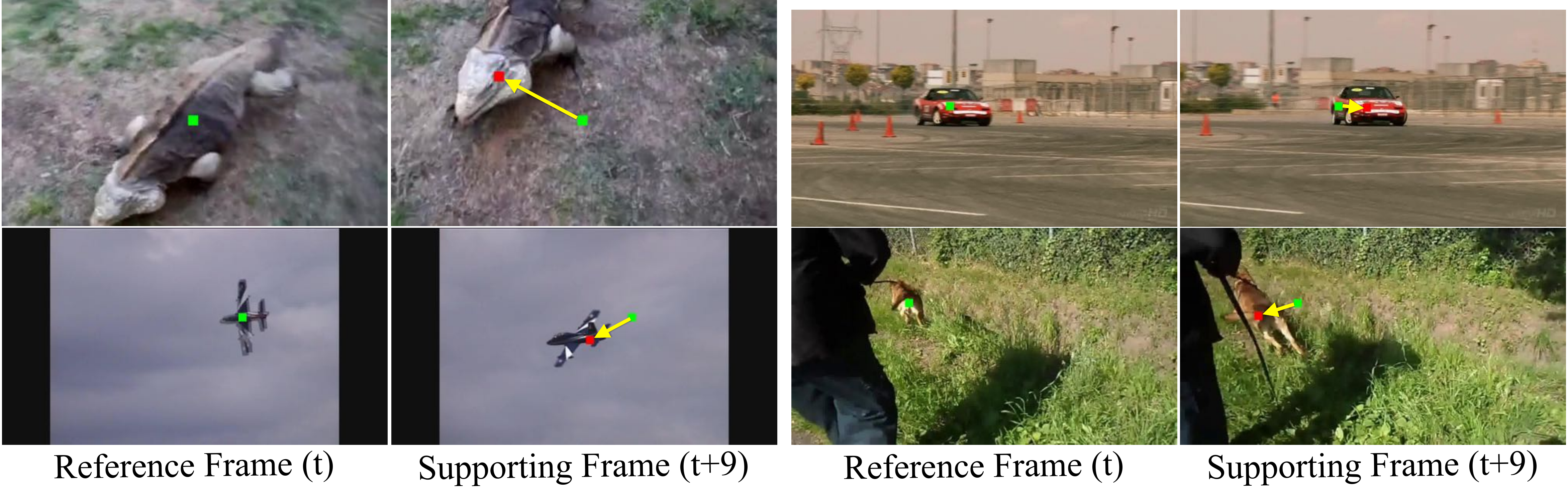}
\end{center}
\vspace{-0.4cm}
  \caption{An illustration of our spatiotemporal sampling scheme (zoom-in for a better view). The green square indicates a point in the reference frame, for which we want to compute a new convolutional output. The red square indicates the corresponding point predicted by our STSN in a supporting frame. The yellow arrow illustrates the estimated object motion.  Although our model is trained discriminatively for object detection and {\em not} for tracking or motion estimation, our STSN learns to sample from the supporting frame at locations that coincide almost perfectly with the same object. This allows our method to perform accurate object detection even if objects in the reference frame are blurry or occluded. \vspace{-0.3cm}}
\label{results1_flow_fig}
\end{figure}

  \captionsetup{labelformat=default}
\captionsetup[figure]{skip=10pt}

\begin{figure}[t]
\begin{center}
   \includegraphics[width=1\linewidth]{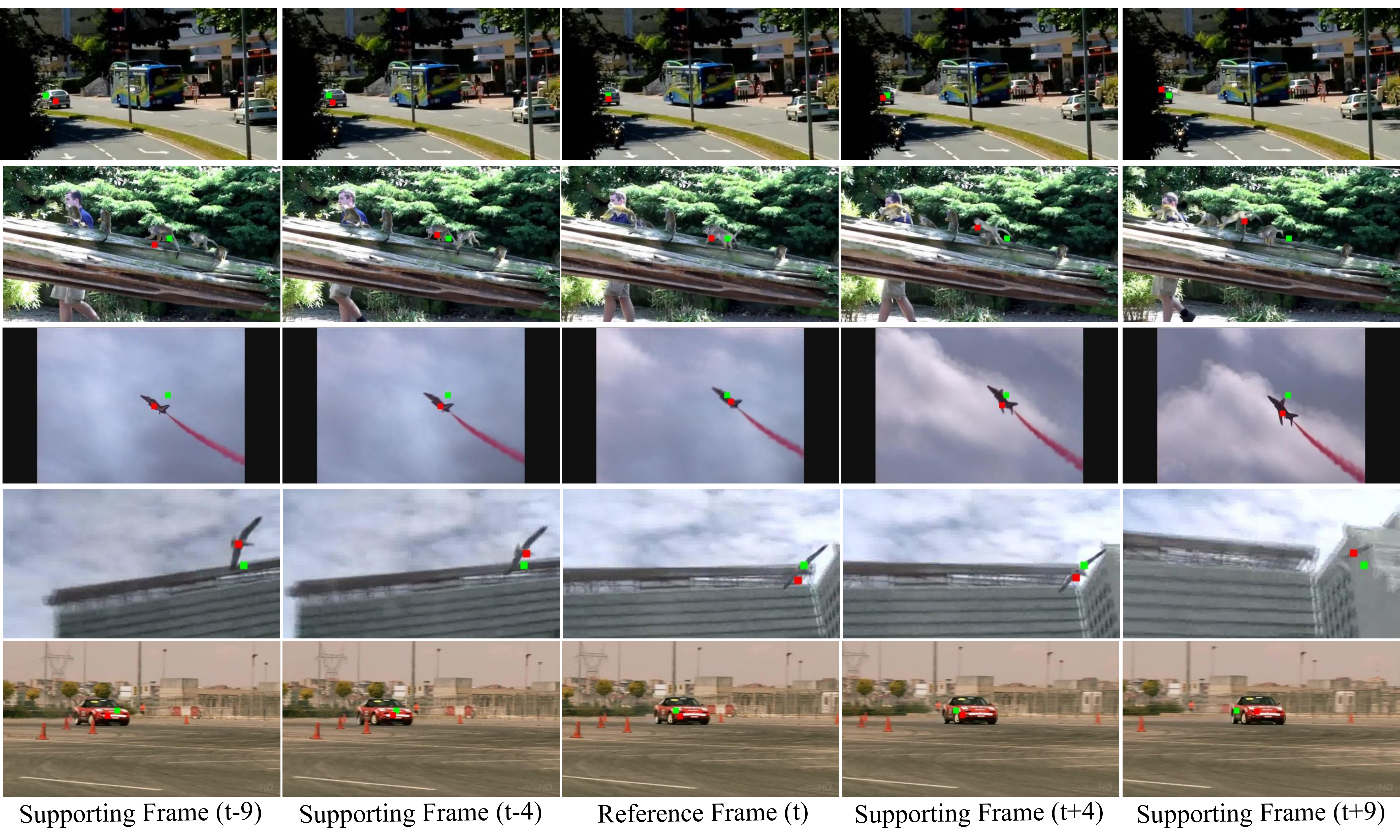}
\end{center}
\vspace{-0.4cm}
  \caption{An illustration of using our spatiotemporal sampling scheme in action. The green square indicates a fixed object location in the reference frame. The red square depicts a location in a supporting frame, from which relevant features are sampled. Even without optical flow supervision, our STSN learns to track these objects in video. In our supplementary material, we include more of such examples in the video format.\vspace{-0.5cm}}
\label{results1_track_fig}
\end{figure}


\subsection{Qualitative Results}

To understand how our STSN exploits temporal information from a given video, we visualize  in Figure~\ref{results1_flow_fig}, the average offsets predicted by the STSN sampling block. These offsets are used by the STSN to decide, which object-level information from the supporting frame should be used to detect an object in the reference frame. The green square in the reference frame depicts a pixel, for which we want to compute a convolution output. The red square in the supporting frame represents an average offset, which is used to determine which feature points from the supporting frame should be sampled. The yellow arrow indicates object's motion between the reference frame and the supporting frame. Note that despite a relatively large motion between the reference and the supporting frames, our STSN samples features from the supporting frame right around the center of the object, which is exactly what we want. Such spatiotemporal sampling allows us to detect objects even if they appear blurry or occluded in the reference frame. 

In addition, based on the results in Figure~\ref{results1_flow_fig}, we observe that even without an explicit optical flow supervision, our STSN learns to accurately capture the motion of the objects, which is another appealing property of our model. In fact, in Figure~\ref{results1_track_fig}, we illustrate several examples of using our STSN to track objects in a given video. From Figure~\ref{results1_track_fig}, we observe that despite a relatively large motion in each sequence, our STSN accurately samples features around objects in every supporting frame. Such results indicate that we may be able to use our sampling mechanism for discriminative object tracking. In fact, we note that the commonly used dense optical flow methods are often redundant because most applications do not require flow prediction for every single pixel. In comparison, we point out that our STSN captures a more discriminative form of motion, which is learned to exclusively benefit a video object detection task. In our supplementary material, we include more of such results in the video form.

\begin{figure}[t]
\begin{center}
   \includegraphics[width=1\linewidth]{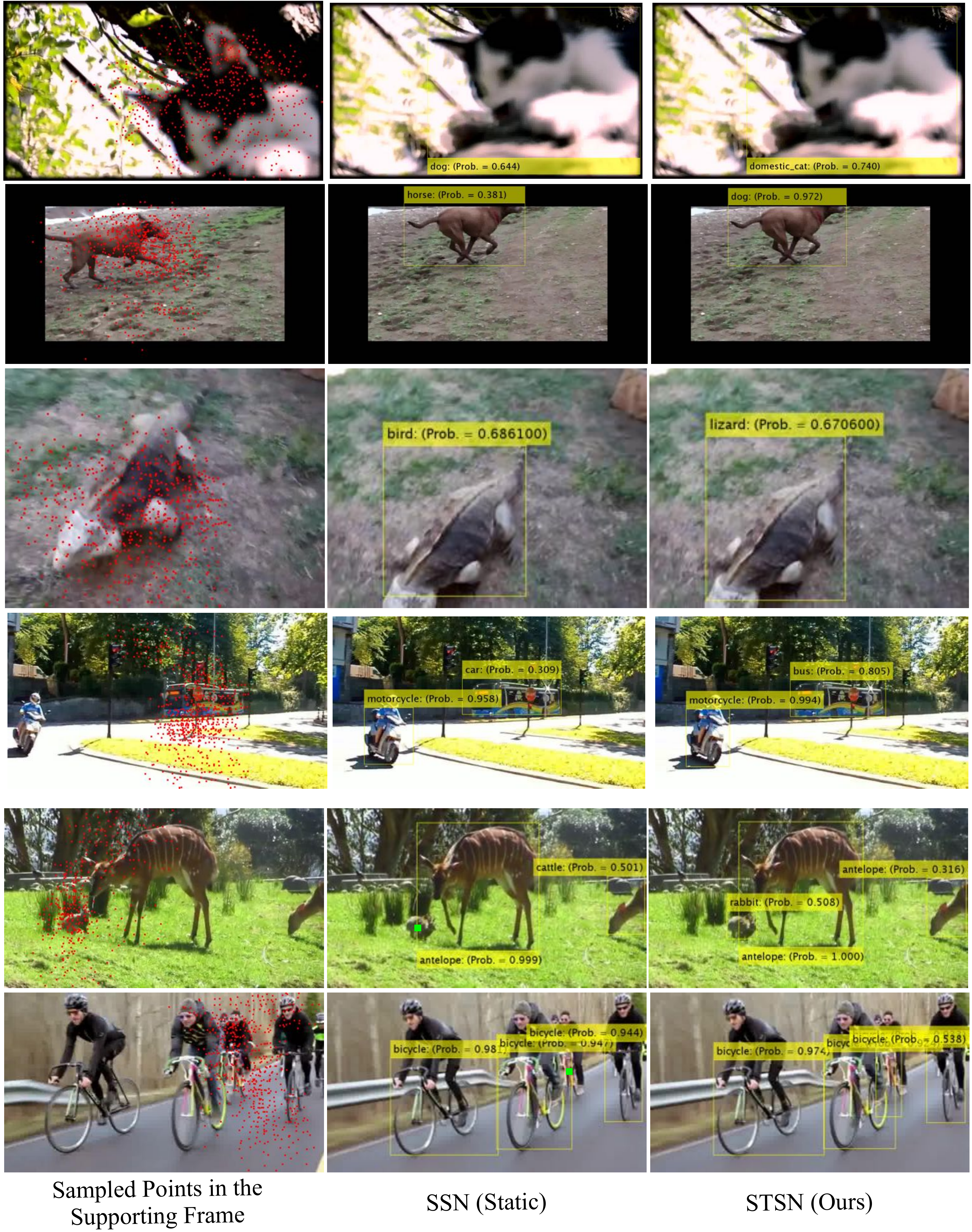}
\end{center}
\vspace{-0.4cm}
    \caption{A figure illustrating object detection examples where our spatiotemporal sampling mechanism helps STSN to correct the mistakes made by a static SSN baseline (please zoom-in to see the class predictions and their probabilities). These mistakes typically occur due to occlusions, blurriness, etc. STSN fixes these errors by using relevant object level information from supporting frames. In Column $1$ we illustrate the points in the supporting frame that STSN considers relevant when computing the output for a point denoted by the green square in Column 2. \vspace{-0.5cm}}
\label{results1_det_fig}
\end{figure}

In Figure~\ref{results1_det_fig}, we also illustrate object detections of the static SSN baseline, and those of our full STSN model (zoom-in to see the probabilities and class predictions). In all of these cases, we observe that incorporating temporal information helps STSN to correct the mistakes made by the static baseline. For instance, in the third row of Figure~\ref{results1_det_fig}, a static SSN baseline incorrectly labels an object in the reference frame as a bird, which happens due to the occluded head of the lizard. However, STSN fixes this mistake by looking at the supporting frames, and by sampling around the lizard body and its head (See Row 3, Column 1 in Figure~\ref{results1_det_fig}). Furthermore, in the last row, a static SSN baseline fails to detect one of the bicycles because it is occluded in the reference frame. STSN fixes this error, by sampling around the missed bicycle in the supporting frame where the bicycle is more clearly visible. Similar behavior also occurs in other cases where STSN successfully resolves occlusion and blurriness issues. 

  \captionsetup{labelformat=default}
\captionsetup[figure]{skip=10pt}

\section{Conclusion}

In this work, we introduced the Spatiotemporal Sampling Network (STSN) which is a new architecture for object detection in video. Compared to the state-of-the-art FGFA~\cite{zhu17fgfa} method, our model involves a simpler design, it does not require optical flow computation and it produces higher video object detection accuracy. Our model is fully differentiable, and unlike prior video object detection methods, it does not necessitate optical flow training data. This renders our model easy to train end-to-end. Our future work will include experimenting with more complex design of spatiotemporal sampling blocks. 

\section{Acknowledgements}

This work was funded in part by NSF award CNS-120552. We gratefully acknowledge NVIDIA and Facebook for the donation of GPUs used for portions of this work.


%
%
%
%
%
%
%
%
%
%
%
%

\clearpage

\bibliographystyle{splncs}
\bibliography{gb_bibliography}

%
%
%
%
\end{document}